\documentclass[10pt,twocolumn,letterpaper]{article}

\usepackage[final]{mystyle}      %
\usepackage{enumitem}

\usepackage{graphicx}
\usepackage{amsmath}
\usepackage{amssymb}
\usepackage{booktabs}
\usepackage{multicol}
\usepackage{multirow}
\usepackage{pifont}
\usepackage{nicefrac}       
\usepackage{makecell}
\usepackage{microtype}

\usepackage[pagebackref,breaklinks,colorlinks]{hyperref}
\usepackage{algorithm}
\usepackage{algorithmic}

\usepackage[capitalize]{cleveref}
\crefname{section}{Sec.}{Secs.}
\Crefname{section}{Section}{Sections}
\Crefname{table}{Table}{Tables}
\crefname{table}{Tab.}{Tabs.}

\begin{document}

\title{Fast Point Cloud Generation with Straight Flows}

\author
{
Lemeng Wu$^{2}\thanks{Work done during an internship at Meta Reality Labs. Correspondence to: Lemeng Wu (lmwu@cs.utexas.edu), Dilin Wang (wdilin@meta.com).}$,
Dilin Wang$^1$,
Chengyue Gong$^2$,
Xingchao Liu$^2$,
Yunyang Xiong$^1$,
Rakesh Ranjan$^1$,\\
Raghuraman Krishnamoorthi$^1$,
Vikas Chandra$^1$,
Qiang Liu$^2$\\
$^1$Meta Reality Labs, $^2$University of Texas at Austin\\
}

\newcommand{\qq}[1]{\textcolor{red}{ql: #1}}
\maketitle

\begin{abstract}
Diffusion models have emerged as a powerful tool for point cloud generation. A key component that drives the impressive performance for generating high-quality samples from noise is iteratively denoise for thousands of steps. While beneficial, the complexity of learning steps has limited its applications to many 3D real-world. To address this limitation, we propose Point Straight Flow (PSF), a model that exhibits impressive performance using one step. Our idea is based on the reformulation of the standard diffusion model, which optimizes the curvy learning trajectory into a straight path. Further, we develop a distillation strategy to shorten the straight path into one step without a performance loss, enabling applications to 3D real-world with latency constraints. We perform evaluations on multiple 3D tasks and find that our PSF performs comparably to the standard diffusion model, outperforming other efficient 3D point cloud generation methods. On real-world applications such as point cloud completion and training-free text-guided generation in a low-latency setup, PSF performs favorably. 

\end{abstract}


\section{Introduction}
3D point cloud generation has many real-world applications across vision and robotics, including self-driving and virtual reality. 
A lot of efforts have been devoted to realistic 3D point cloud generation, such as VAE~\cite{kim2021setvae}, GAN~\cite{achlioptas2018learning,shu20193d}, Normalizing Flow~\cite{yang2019pointflow,kim2020softflow,klokov2020discrete} and score-based method~\cite{luo2021diffusion,zhou20213d,cai2020learning,zeng2022lion}, and diffusion model~\cite{zeng2022lion,luo2021diffusion,zhou20213d}. 
Among them, diffusion models gain increasing popularity for generating realistic and diverse shapes by separating the distribution map learning from a noise distribution to a meaningful shape distribution into thousands of steps. 


\begin{figure*}[!bpht]
   \centering
    \includegraphics[width=0.98\textwidth]{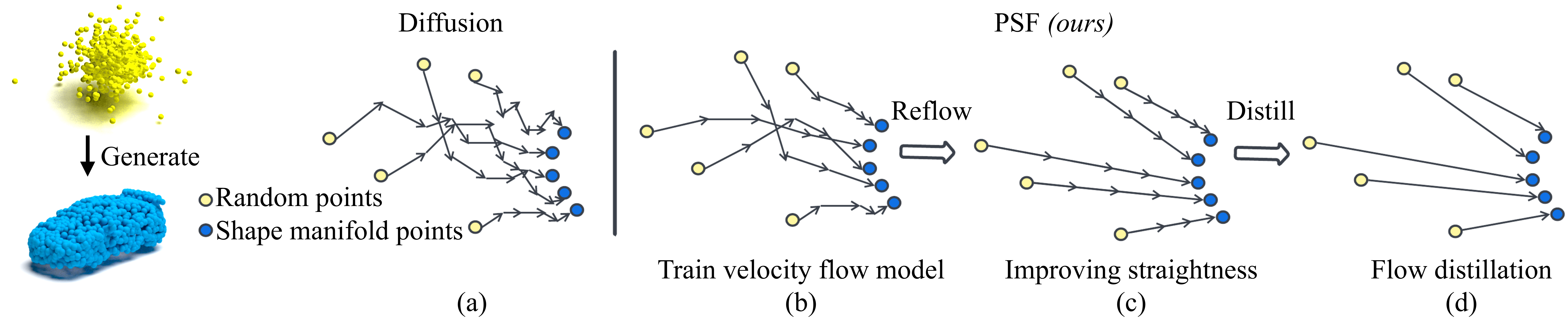}

    \caption{Trajectories during the generate process for the point cloud generation. (a) The SDE (PVD) trajectory involves random noise in each step and thus gives a curvy trajectory. (b) The 
    PSF initial flow model removes the random noise term and gets a simulation procedure trajectories with smaller transport cost. 
    (c) By utilizing the \textit{reflow} process on the initial flow model, we reduce the transport cost. As a result, the trajectories are becoming straightened and easy to simulate with one step. (d) The straight path leads to a small time-discrimination error during the simulation, which makes the model easy to distill into one-step. }

    \label{fig:traj}
\end{figure*}


Despite the foregoing advantages, the transport trajectory learning from a noise distribution to a meaningful shape distribution also turns out to be a major efficiency bottleneck during inference since a diffusion model requires thousands of generative steps to produce high-quality and diverse shapes~\cite{ho2020denoising,zhou20213d,song2020denoising}. As a result, it leads to high computation costs for generating meaningful point cloud in practice. Notice that the learning transport trajectory follows the simulation process of solving stochastic differentiable equation~(SDE). A trained neural SDE can have different distribution mappings at each step, which makes the acceleration challenging even with an advanced ordinary differentiable equation (ODE) solver.


To address this challenge, several recent works have proposed strategies that avoid using thousands of steps for the meaningful 3D point cloud generation. For example, ~\cite{salimans2022progressive,luhman2021knowledge} suggest distilling the high-quality 3D point cloud generator, DDIM model~\cite{song2020denoising}, into a few-step or one-step generator. While the computation cost is reduced by applying distillation to shorten the DDIM trajectory into one-step or few-step generator. 
 The distillation process learns a direct mapping between the initial state and the final state of DDIM, which needs to compress hundreds of irregular steps into one-step. Empirically it leads to an obvious performance drop. Further, these distillation techniques are mainly developed for generating images with the grid structure, which is unsuitable for applying to point cloud generation since the point cloud is an unordered set of points with irregular structures.

In this paper, we propose using one-step to generate 3D point clouds. Our method, Point Straight Flow (PSF), learns a straight generative transport trajectory from a noisy point cloud to a desired 3D shape for acceleration. This is achieved by passing the neural flow model once to estimate the transport trajectory. Specifically, we first formulate an ODE transport flow as the initial 3D generator with a simpler trajectory compared with the diffusion model formulated in SDE. Then we optimize the transport flow cost for the initial flow model to significantly straighten the learning trajectory while maintaining the model's performance by adopting the ideas from recent works ~\cite{liu2022flow,lipman2022flow}. This leads to a straight flow by optimizing the curvy learning trajectory into a straight path. Lastly, with the straight transport trajectory, we further design a distillation technique to shorten the path into one-step for 3D point cloud generation. 

To evaluate our method, we undertake an extensive set of experiments on 3D point cloud tasks. We first verify that our one-step PSF can generate high-quality point cloud shapes, performing favorably relative to the diffusion-based point cloud generator PVD~\cite{zhou20213d} with a more than $700\times$ faster sampling on the unconditional 3D shape generation task. Further, we demonstrate it is highly important to learn straight generative transport trajectory for faster sampling by comparing distillation baselines, including DDIM that are difficult to generate shapes even with many more generative steps. Finally, we perform evaluations on 3D real-world applications, including point cloud completion and training-free text-guided generation, to show the flexibility of our one-step PSF generator. 

%

\begin{itemize}[noitemsep,topsep=0pt]
    \item For the first time, we demonstrate that neural flow trained with one step can generate high-quality 3D point clouds by applying distillation strategy.
    \item We propose a novel 3D point cloud generative model, Point Straight Flow (PSF). Our proposed PSF optimizes the curvy transport trajectory between noisy samples and meaningful point cloud shapes as a straight path.  
    \item We show our PSF can generate 3D shapes with high-efficiency on standard benchmarks such as unconditional shape generation and point cloud completion. We also successfully extend PSF to real-world 3D applications including large-scene completion and training-free text-guided generation.
\end{itemize}


\section{Related Work}

\subsection{Generative model with transport flow}
The generative model can be treated as transporting a distribution to another. Previous works like VAE~\cite{kingma2013auto} and GAN~\cite{goodfellow2014generative} build this transport in one-step using neural networks. However, as the data capacity and complexity increase, this one-step mapping takes a lot of effort to train. Recently, the community tries to relieve this issue by decomposing the one-step mapping into several steps in ODE~\cite{chen2018neural,papamakarios2021normalizing,song2020denoising} or SDE~\cite{song2020score,song2019generative,tzen2019theoretical,ho2020denoising} fashion. Among these works, denoising diffusion probabilistic models (DDPM)~\cite{ho2020denoising} demonstrate the power and flexibility to generate high-quality samples on large-scale image benchmarks and other domains ~\cite{nichol2021improved,dhariwal2021diffusion,ramesh2022hierarchical,rombach2022high,ho2022video,yang2022diffusion}, which makes the diffusion model become a mainstream to learn the transport flow.

\subsection{Fast sampling for transport model}

Despite the huge success of DDPM, a major issue of this model is that it requires thousands of steps to generate high-quality desired samples. Previous works propose multiple strategies to decrease the simulation steps and accelerate the DDPM to learn transport process. DDIM~\cite{song2020denoising} formulates the sampling trajectory process as an ODE. FastDPM~\cite{kong2021fast} bridges the connection between the discrete and continuous time step. These two methods can help reduce the learning trajectory to hundreds of steps. Beyond this, to further compress the generation process to few-step simulation,  ~\cite{luhman2021knowledge,salimans2022progressive} apply  knowledge distillation to learn a few mappings to recover the multiple DDIM steps. However, these methods are hard to maintain a good performance with a single step or even more steps. Recent ODE transport works~\cite{liu2022flow,liu2022rectified,lipman2022flow,albergo2022building} try to build the ODE transport with a reduced cost for a faster simulation compared with DDIM. Motivated by the above ideas, we optimize the transport cost first to learn  the transport trajectory with one step in our PSF model.

\subsection{3D generative model }
Generating 3D objects enables shape creation and completion for various applications. Previous literature focus on using VAE~\cite{brock2016generative,kim2021setvae}, GAN~\cite{3dgan,achlioptas2018learning,gao2022get3d,li2021sp} and Normalizing Flow~\cite{yang2019pointflow,kim2020softflow,klokov2020discrete} to generate the 3D object in point cloud, mesh and voxel representations.  Recently, more dedicated shapes can be generated using the score-based or diffusion models~\cite{cai2020learning, luo2021diffusion,zhou20213d,zeng2022lion,zheng2022neural,lyu2021conditional}. 
While the diffusion model in the point cloud generation obtains state-of-the-art results, it is computationally expensive to generate even one high-quality shape. This makes the diffusion generative model difficult to apply to real-world 3D applications. Our PSF boosts the speed under 0.1s while maintaining the performance compared with diffusion-based methods.

\section{Point Straight Flow}
We now introduce our method, Point Straight Flow (PSF),  one-step transport flow for 3D point cloud generation. 
Generating point clouds with transport flow can be viewed as transporting noise point clouds to target data point clouds by following a learned trajectory.
To consolidate the transport steps as few as one, 
we propose a three-stage training pipeline as illustrated in Figure~\ref{fig:traj}:
1) Our first step is to learn a neural velocity flow network. We construct an ODE process with the shortest transport path and utilize a neural network to fit this process.
At the end of this step, one can start with random Gaussian noises and then iteratively apply the ODE process from the learned network to generate samples.
However, the training object does not exactly match the generative process. Thus, the trajectory could still be curvy. 
2) Our second step is to optimize the trajectory's straightness learned at step 1 via \emph{reflow} adapted from ~\cite{liu2022flow}. This \emph{reflow} stage encourages the velocity network to flow straightly while performing sample  generation. This allows us to combine multiple updates into one easily.
3) We further distill the neural network with the objective such that one step 
with a fixed large step size (Figure~\ref{fig:traj} (c)) makes the same update as the iteration of multiple small updates as in Figure~\ref{fig:traj} (b). 
\paragraph{Training initial velocity flow network}
Our goal is to build a transport flow to push the point clouds from 
a Gaussian distribution to the point cloud data distribution. 

Specifically, assume the point cloud in the xyz-coordinates and denote 
$X_0\in \mathbb{R}^{M \times 3}$ as Gaussian noise and $X_1\in \mathbb{R}^{M \times 3}$ as real data samples.
We denote $v_\theta$ as the velocity field network by the following ODE process,
\begin{equation}
    \underbrace{\mathrm{d} X_t}_\text{drift} = \underbrace{v_\theta(X_t, t)}_\text{velocity} \underbrace{\mathrm{d}t}_\text{time interval}, \quad \text{with}~~t\in[0, 1].
    \label{eq:ode}
\end{equation}
Here $X_t$ is the intermediate point cloud states at time $t$ and the velocity filed $v_\theta: \mathbb{R}^{M \times 3} \rightarrow \mathbb{R}^{M \times 3}$ is a neural network with $\theta$ as its parameters. 



Given the intermediate point cloud $X_t$, $v_\theta$ defines a velocity field that moves $X_t$ further towards true data $X_1$. Intuitively, the optimal direction at any time $t$ is $X_1 - X_0$. Thus, we can encourage our velocity field to directly follow the optimal ODE process $\mathrm{d}X_t = (X_1-X_0)\mathrm{d}t$ by optimizing
\begin{align}
\begin{split}
    & \min_\theta \int_0^1 \mathbb{E} \bigg [\|(v_\theta(X_t, t) - (X_1 - X_0)\|^2 \bigg]\mathrm{d}t,  
    \\ 
    & \mathrm{where}~~ X_t = t X_1 + (1-t)X_0 \quad t\in[0, 1]. 
    \label{eq:main}
    \end{split}
\end{align}
Empirically, we do not optimize the loss in Eqn.~\ref{eq:main} with the integration on $t \in [0, 1]$ directly. Instead, for each data sample $X_1$, we randomly draw a $X_0$ from Gaussian noise, a $t$ from [0, 1] and minimize the following equivalent loss, 
\begin{align}
\min_\theta \mathbb{E} \bigg[\|(v_\theta(X_t, t) - (X_1 - X_0)\|^2   \bigg].
~~~ t \sim \mathcal{U}(0, 1).
\label{eq:straightness}
\end{align}

After the neural velocity field is well-trained, 
 samples can be generated by discretizing the ODE process with Euler solver in Eqn.~\ref{eq:ode} into $N$ steps (e.g., $N=1000$) as the following,
\begin{align}
    X'_{(\hat{t}+1)/N} \longleftarrow X'_{\hat{t}/N} + \frac{1}{N} ~v_\theta(X'_{\hat{t}/N}, \frac{\hat{t}}{N}),
    \label{eq:sample}
\end{align}
the integer time step $\hat{t}$ is defined as $\hat{t} \in \{0,1,\cdots,N-1\}$. Here $X'_{1}$ denotes our generated samples and $X'_0 = X_0$.
Intuitively, the solver will be more accurate with a large $N$.

\begin{figure}[tb]
    \centering
    \includegraphics[width=0.48\textwidth]{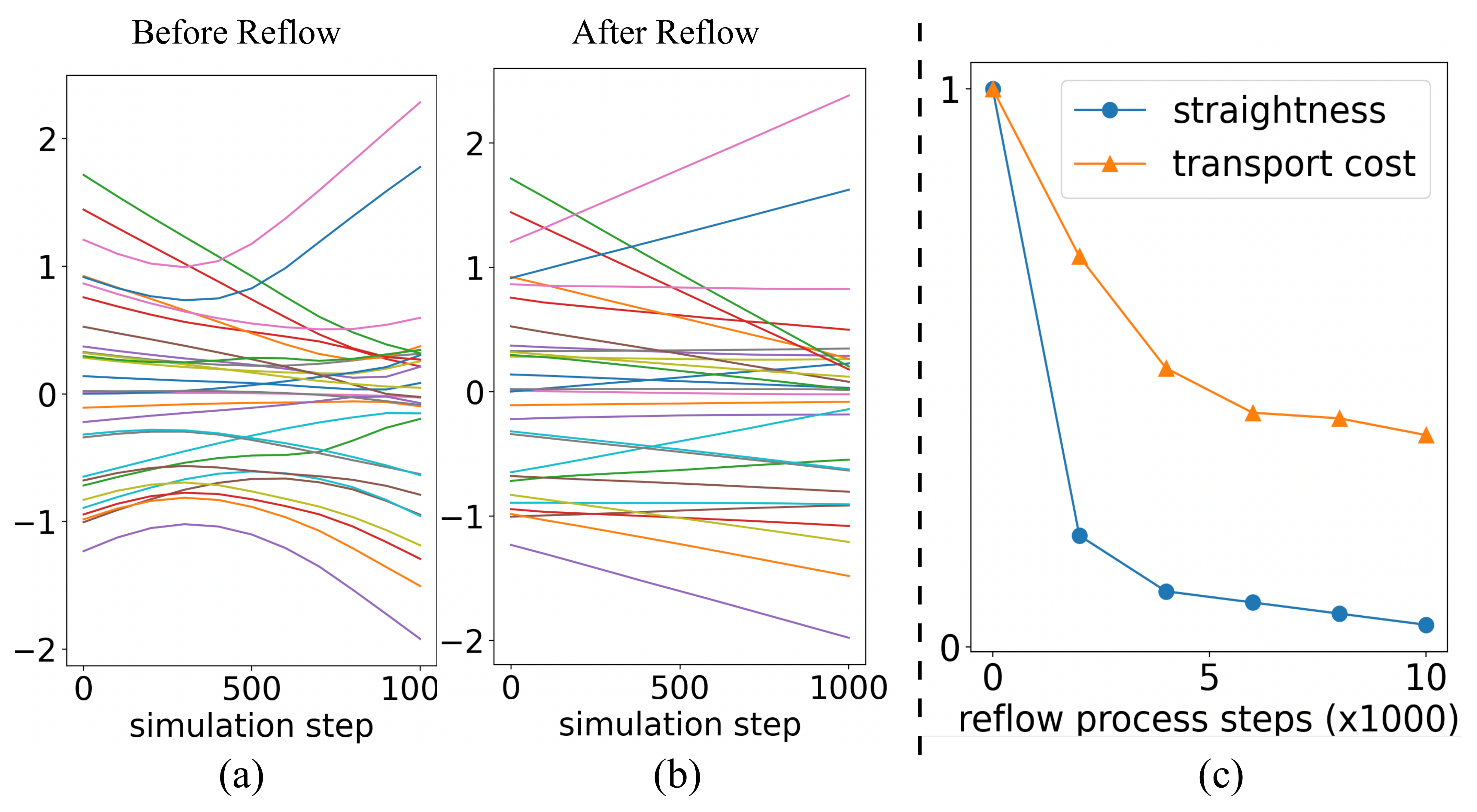}
    \caption{We visualize the impact of \textit{reflow}. (a) and (b) shows the generation trajectory before and after \textit{reflow}. (c) shows the transport and straightness changes during the \textit{reflow} finetune. We use value 1 to represent the initial value before the \textit{reflow} and 0 as the ideal value that represents the optimal transport or strict straightness.}
  
    \label{fig:straight}
\end{figure}

\paragraph{Improving straightness of the flow}
In the previous stage, network $v_\theta$ is trained on data pairs $(X_0, X_1)$,
with $X_1$ as ground truth data points.
We empirically found the trajectory learned in this way is still curvy during the generative process. Therefore, a large number $N$ is necessary for accurate approximation of Eqn.~\ref{eq:ode} with Euler solver as in Eqn.~\ref{eq:sample}. 

Specifically, in Figure~\ref{fig:straight} (a), 
we show an example of unconditional point cloud generation and plot the movement trajectories of 30 randomly sampled points with a randomly picked dimension on x/y/z.
The update of each point is guided by Eqn.~\ref{eq:sample} with $N=1000$.
See Section~\ref{exp:uncondition} for more detailed discussions on experiment settings.

Intuitively, one would prefer a straight trajectory, which means a smaller $N$ needs to give a similar output as a bigger $N$, namely, generating samples with fewer updates.
To this end, we sample a set of Gaussian noise $X'_0$ and generate its corresponding samples $X'_1$  following Eqn.~\ref{eq:sample} with $v_{\theta}$ fixed.
After that, one can finetune the $v_\theta$ using the sampled pairs $(X'_0, X'_1)$ follow Eqn.~\ref{eq:main} by replacing the $(X_0, X_1)$. 
This is referred to as the \emph{reflow} procedure in \cite{liu2022flow}. Theoretically, the sampled fixed pairs $(X'_0, X'_1)$ always provide a lower transport cost than pairs $(X_0, X_1)$. 
Thus training on the $(X'_0, X'_1)$  can reduce the transport cost and straighten the trajectory. 

In Figure~~\ref{fig:straight} (b), we show the resulting trajectories after applying \emph{reflow} procedure by still taking $N=1000$.
As we can see from this figure, the trajectories are nearly  straight for all the points.
In Figure~\ref{fig:straight} (c), we quantitatively show the change of the transport cost and straightness after the \textit{reflow} procedure. The transport cost is defined as $l_2$ transport cost and 
the straightness of trajectories can be written as the following, 
\begin{equation}
    \text{Straightness} =  \frac{1}{N} \sum_{\hat{t}=0}^{N-1} \bigg[\parallel(X'_1 - X'_0) ~-~ v_\theta(X_{\hat{t}/N}, \hat{t}/N)\parallel^2\bigg].
    \label{eq:straight}
\end{equation}
A straight trajectory between $X'_0$ and $X'_1$ implies a constant velocity  $v_\theta(X_{\hat{t}/N}, \hat{t}/N) = X'_1 - X'_0$ at every time $\hat{t}$ step. And Eqn.~\ref{eq:straight} equals 0 in this case.
\paragraph{Flow distillation}
Our observation above indicates the possibility to approximate  Eqn.~\ref{eq:ode} with only one discretization step if the trajectories are straight, 
\begin{equation} 
    X'_1 = X'_0 + v_\theta(X'_0, 0).
    \label{eq:one_step}
\end{equation}
And our goal is to make sure the update in Eqn.~\ref{eq:one_step} leads to samples that have similar quality as samples generated by Eqn.~\ref{eq:sample} with a large $N$.
See Figure~\ref{fig:traj} (c) for an illustration.




For this purpose, we introduce a third \emph{distillation stage} to summarize the refined ODE process with $v_\theta$ in the previous stage.
In particular, we construct the distillation objective as follows,
\begin{equation}
    \min_\theta \mathbb{E}\bigg[\mathrm{Dist}(\underbrace{X'_0 + v_\theta(X'_0, 0)}_\text{one step}, \underbrace{X'_1}_\text{$N$ step})\bigg],
    \label{eq:distill}
\end{equation}
where $\mathrm{Dist}(\cdot)$ is a loss function that measures the difference between two sets of point clouds. And $X'_1$ is generated with $N=1000$ same as the previous \textit{reflow} step.

In the literature~\cite{luhman2021knowledge,salimans2022progressive}, 
the $\text{Dist}(\cdot)$ is typically constructed as a $\ell_2$ loss.
However, unlike images which impose a strict alignment at each pixel location during construction, 
the permutation invariant property of point clouds makes $\ell_2$ loss less suitable. 
Hence we propose the Chamfer distance as the objective for distillation. 
Specifically, assume $X_i$, $X_j$ as two point clouds, the Chamfer distance is defined as 
\begin{equation}
    \text{CD}(X_i, X_j) = \sum_{p\in X_i} \min_{\hat{p} \in X_j} || p - \hat{p} ||_2 + \sum_{\hat{p} \in X_j} \min_{p \in X_i} || p - \hat{p} ||_2.
\end{equation}

\paragraph{Summary} We summarize the overall algorithm in Algorithm~\ref{alg:main}, and refer readers to the Appendix for the training and sampling pseudo-code. 
Overall, our algorithm gives a one-step point cloud generation approach. 
After these three training stages, one can sample a point cloud in one step starting from a random noise by following Eqn.~\ref{eq:one_step}.

\begin{algorithm}[!bpht]
\caption{Point Straight Flow} \label{alg:main}
\begin{algorithmic}
\STATE \textbf{Input}: Point cloud dataset $\mathcal{D}$, a neural velocity field $v_\theta$ with parameter $\theta$.
\STATE \textbf{1. Training initial velocity flow model}: randomly sample  $X_0 \sim \mathcal{N}(\textbf{0}, \textbf{I})$  and  $X_1 \sim \mathcal{D}$, and train  $v_\theta$ follows the objective function Eqn.~\ref{eq:straightness} to convergence.
\STATE \textbf{2. Improving straightness via \textit{reflow}}: Sample a set of point cloud pairs, with $X'_0 \sim \mathcal{N}(\textbf{0}, \textbf{I})$, and $X'_1$ generated with Eqn.~\ref{eq:sample}. 
Use $(X'_0, X'_1)$ as training data to finetune $v_\theta$ by still minimizing Eqn.~\ref{eq:straightness}.
\\
\STATE \textbf{3. Flow distillation}: Use pairs $(X'_0, X'_1)$ to finetune $v_\theta$ with the distillation loss Eqn.~\ref{eq:distill} into one-step generator as the final PSF model.
\STATE \textbf{Sampling (output)}: Randomly sample from $X'_0 \sim \mathcal{N}(\textbf{0}, \textbf{I})$, and output the desired point cloud $X'_1$ with $X'_1 = X'_0 + v_\theta(X'_0, 0)$.
\end{algorithmic}
\end{algorithm}

\section{Experiment}
\begin{figure*}[!bhpt]
    \centering
    \includegraphics[width=0.98\textwidth]{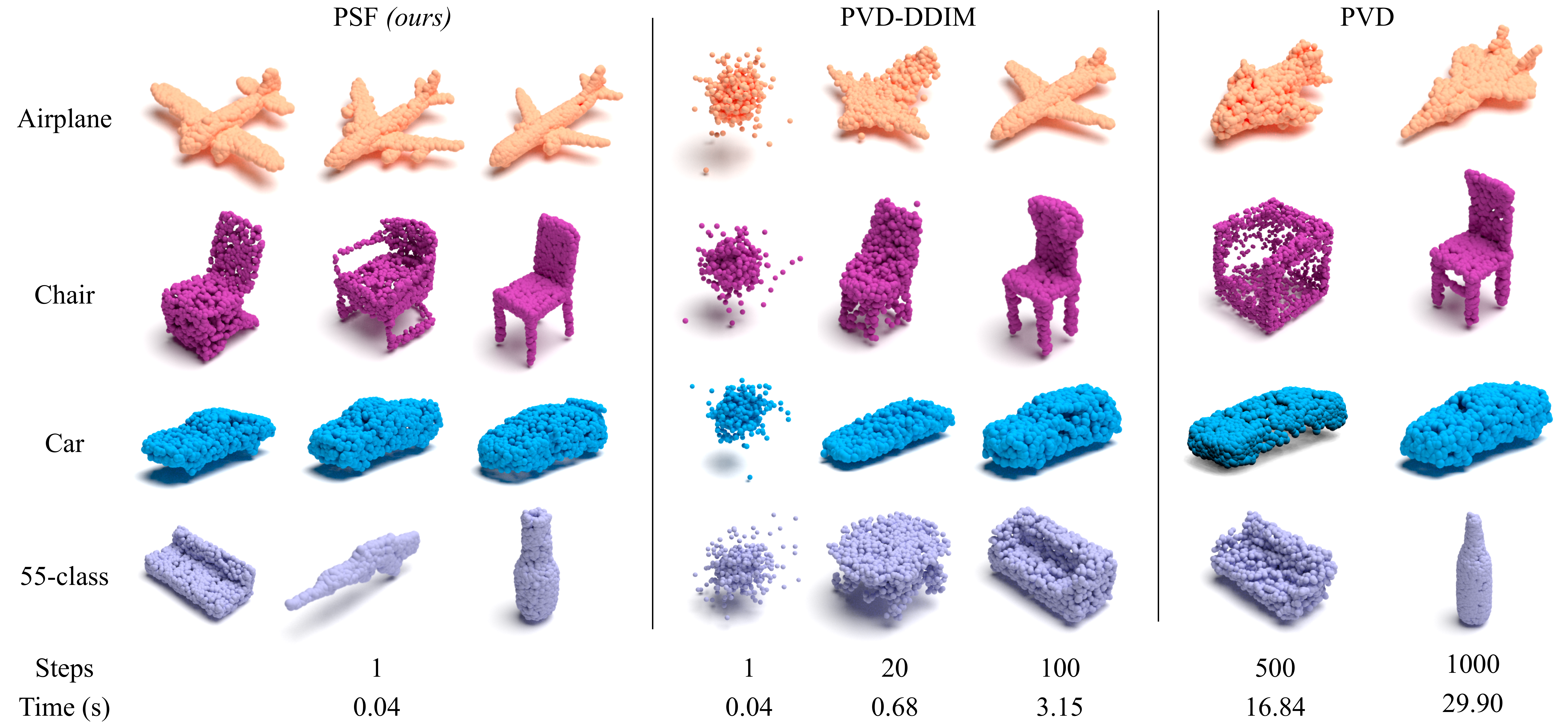}
    \caption{Visualization for the unconditional generation for Airplane, Chair, Car and 55-class. We show PSF can generate high-quality point cloud in one-step with only 0.04 seconds. }
    \vspace{-0.5em}
    \label{fig:comparision}
\end{figure*}

We empirically demonstrate the effectiveness and efficiency of our method on 
three tasks, including unconditional point cloud generation, training-free text-guided point cloud generation and point cloud completion.
Compared with the prior-art diffusion-based point cloud generalization method PVD~\cite{zhou20213d}, our method reduces the GPU inference time by $75\times$ while still producing realistic samples at a similar quality level.
\paragraph{General training settings}
\label{sec:exp}
We adopt the PVCNN~\cite{pvcnn} styled U-Net~\cite{ronneberger2015u} proposed in PVD~\cite{zhou20213d} as our velocity filed model $v_{\theta}$. 
As we summarize in Algorithm~\ref{alg:main}, there are three phases, 
including 1) training the velocity flow model, 2) improving straightness via \emph{reflow}, and 3) flow distillation. 
For step 1), we mainly follow the setting in DDPM~\cite{ho2020denoising} and use a batch size of 256 and a learning rate of $2e^{-4}$. We train the model for 200k steps and apply an exponential moving average (EMA) at a rate of 0.9999. 
For step 2), we randomly sample 50k data pairs $(X'_0, X'_1)$ using the pretrained network $v_{\theta}$ and fine-tune it for 10k steps by minimizing Eqn.~\ref{eq:main}. 
Here $X'_1$ is sampled by setting $N=1000$ in Eqn.~\ref{eq:sample} for the best quality. 
We use a fixed learning rate of $2e^{-5}$ in this step. 
For step 3),
we use the samples $(X'_0, X'_1)$ generated in step 2) and
finetune $v_\theta$ for another 10k steps with learning rate $2e^{-5}$. 


\subsection{Unconditional point cloud generation}
\label{exp:uncondition}
We apply our method PSF for unconditional 3D point cloud generation. Compared with existing solutions, our method runs as fast as
one-shot VAE-based (e.g.,SetVAE~\cite{kim2021setvae}) or GAN-based (e.g., 1-GAN~\cite{achlioptas2018learning}) approaches, 
in the meantime, our method generates samples as high quality as computation expensive diffusion-based methods like PVD. 
Additionally, 
we also compare PSF with a fast-sampling method DDIM~\cite{song2020denoising} combined with PVD, we denote this approach as PVD-DDIM.  Compared with PVD-DDIM, our method achieves much better sample quality as well as sampling efficiency.
\vspace{-0.5em}
\paragraph{Settings}
We use the same training and testing data split by following 
 PointFlow~\cite{yang2019pointflow} and PVD~\cite{zhou20213d}. 
We choose one nearest neighbor accuracy (1-NNA) with Chamfer Distance~(CD) and Earth Movement Distance~(EMD) as our metrics to measure the sample quality per the suggestion in PVD. Please refer Appendix for other common metrics, including Matching Distance (MMD) and Coverage Score (COV).
We test the sampling time of our method and baselines on Nvidia RTX 3090 GPUs with a batch size of 1. All the time comparisons are averaged on 50 random trials. 
For the PVD-DDIM setup, we grid search the time step $N$ in $\{1, 20, 50, 100, 500, 1000\}$ and report the $N=100$, which is the smallest step with 1-NNA performance degrate no larger than 10 $\%$ compared with PVD. 

\begin{table*}[]
    \centering
        
     \begin{tabular}{l|c|cc|cc|cc}
\Xhline{3\arrayrulewidth} 
      \multirow{2}{*}{Model} & \multirow{2}{*}{Sampling Time (s)} &  \multicolumn{2}{c|}{Airplane} & \multicolumn{2}{c|}{Chair}  & \multicolumn{2}{c}{Car} \\
      \cline{3-8}
      & & CD $\downarrow$ & EMD $\downarrow$ & CD $\downarrow$ & EMD $\downarrow$ & CD $\downarrow$ & EMD $\downarrow$\\
    \Xhline{1.5\arrayrulewidth} 
      1-GAN~\cite{achlioptas2018learning} & \textbf{0.03} & 87.30 & 93.95 & 68.58 & 83.84 & 66.49 & 88.78 \\
       PointFlow~\cite{yang2019pointflow} & 0.27 & 75.68 & 70.74 & 62.84 & 60.57& 58.10 & 56.25 \\
       DPF-Net~\cite{klokov2020discrete} & 0.33 & 75.18 & 65.55 & 62.00 & 58.53& 62.35  &54.48  \\
       SoftFlow~\cite{kim2020softflow} & 0.12 & 76.05 & 65.80 & 59.21 & 60.05& 64.77 & 60.09 \\
       SetVAE~\cite{kim2020softflow} & \textbf{0.03} &75.31 & 77.65 & 58.76 & 61.48 & 59.66 & 61.48 \\
       ShapeGF~\cite{cai2020learning} & 0.34 & 80.00 & 76.17 & 68.96  & 65.48& 63.20 & 56.53 \\
       
        DPM~\cite{luo2021diffusion} & 22.8 & 76.42 & 86.91 & 60.05 & 74.77 & 68.89 & 79.97 \\
       PVD~\cite{zhou20213d} (N=1000) & 29.9 & \textbf{73.82} & \textbf{64.81} & \textbf{56.26} & \textbf{53.32} & \textbf{54.55} & \textbf{53.83} \\
        \Xhline{1.5\arrayrulewidth} 
      
        PVD-DDIM~\cite{song2020denoising} (N=100)  & 3.15 & 76.21 & 69.84 & 61.54 & 57.73 &  60.95 & 59.35 \\
        PSF \textit{(ours)} & \textbf{0.04} & \textbf{71.11} & \textbf{61.09} & \textbf{58.92} & \textbf{54.45}& \textbf{57.19} & \textbf{56.07} \\
 \Xhline{3\arrayrulewidth} 
       
    \end{tabular}
    \caption{\emph{Performance} (1-NNA $\downarrow$) and \emph{Sampling Time} on single class generation. The second block represents the fast simulation methods. We report the smallest step size of PVD and PVD-DDIM 
    , which do not drop performance. 
    The sampling time is calculated when the batch size is one. }
    \vspace{-0.5em}
    \label{tab:main_result}
\end{table*}

\begin{figure*}[!bhpt]
    \centering
    \includegraphics[width=1.0\textwidth]{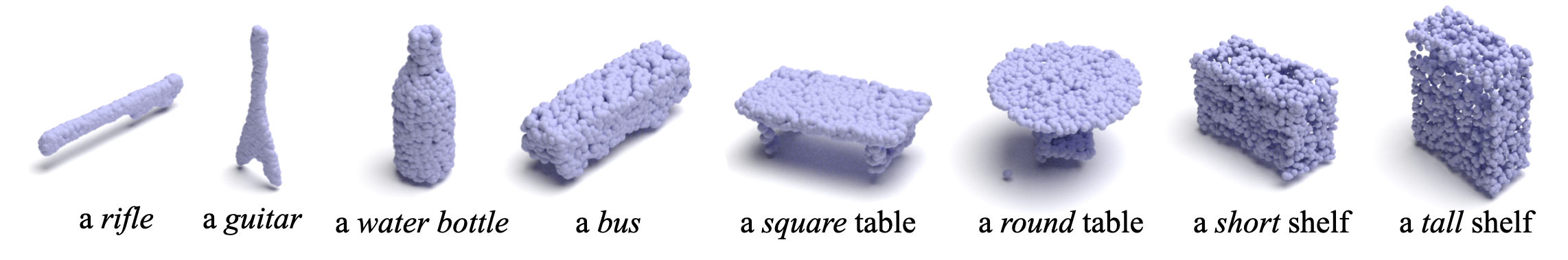}
    \caption{Training-free text-guided point cloud generation using CLIP loss. We show that our 55-class pretrained PSF can generate the correct and high-quality shapes following the text prompt.}
    \label{fig:textguide}
    \vspace{-0.5em}
\end{figure*}

\paragraph{Results}
We first show qualitative comparisons in Figure~\ref{fig:comparision}.
Our method generates samples of similar visual quality compared to the samples from the expensive PVD approach with 1000 steps. 
Compared with PVD-DDIM in 100 steps and PVD in 500 steps,
our PSF produces better samples with clear shapes and boundaries.  

Additionally, we show quantitative comparisons in Table~\ref{tab:main_result}.
Overall, our method PSF achieves similar CD and EMD scores compared with PVD (N=1000) in all three categories. 
The performance is only slightly worse on the Chair and Car, while the visual quality is almost the same as demonstrated in Figure~\ref{fig:comparision}.
Most significantly, the average sampling time of our method is only 0.04s on a modern GPU, which is more than $75\times$ faster compared with PVD-DDIM (N=100) and $700\times$ faster compared with PVD (N=1000). 
We provide the additional comparisons for other time-step setups for PVD, PVD-DDIM and PSF in the ablation study in Section~\ref{sec:ablation}.
\paragraph{Initial state linear interpolation leads to interpretable interpolation for final samples }

\begin{figure*}[ht]
    \centering
    \includegraphics[width=1.0\textwidth]{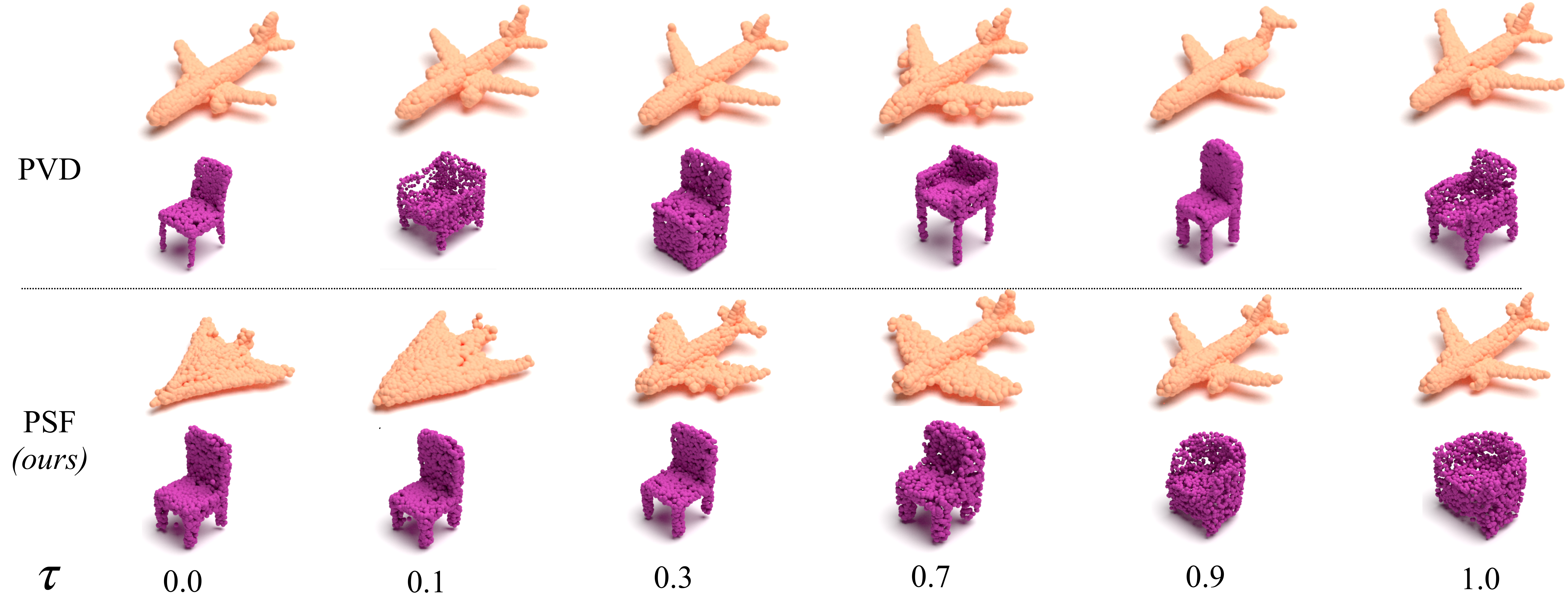}
    \caption{Interpolation of two random picked initialization.
    Our PSF can simulate the interpolated initial status with a continuously changing shape, while PVD simulation the shapes without a relationship. }
    \label{fig:linear}
\end{figure*}

Besides sampling efficiency, 
we show another benefit of our method: the ability to generate interpretable point clouds by starting from interpolated noise due to the straightness of the velocity trajectory. 
Specifically, we first randomly draw initial 
point clouds $\tilde{x}_{0}$ and $\tilde{x}_{1}$ from a Gaussian distribution, 
then we apply the linear interpolation between $\tilde{x}_{0}$ and $\tilde{x}_{1}$ to generate additional Gaussian noise, 
\begin{equation*}
    \tilde{x}_{\tau} = \sqrt{(1 - \tau)} \tilde{x}_{0} + \sqrt{\tau} \tilde{x}_{1} ~~~ \tau\in[0, 1].
\end{equation*}  

We apply both our method and PVD to generate samples by taking $\{\tilde{x}_{\tau}\}$ as the starting points. We present our findings in Figure~\ref{fig:linear}.
As we can see from the Figure, a smooth change in inputs leads to a smooth change between the final generated samples for our method. 
While PVD trained with diffusion algorithms doesn't enjoy this property. 


\subsection{Training-free text-guided shape generation}
\label{sec:clip}
In this section, we extended our method for training-free (i.e., with $v_\theta$ fixed) text-guided point cloud generation. 
Specifically, given a text input, training-free text-guided sample generation can be framed as finding the best initial noise $X_0$ such that the resulting generated sample matches the text prompt semantically. 

We follow the setting in Text2Mesh~\cite{michel2022text2mesh} and FuseDream~\cite{liu2021fusedream} for sample optimization, 
\begin{equation}
    \min_{X_0} \mathbb{E}_{\{\mathrm{Proj}_i\}} \bigg[ S_{\mathrm{clip}}\bigg (\mathrm{Proj}_i \cdot \mathrm{generator}(X_0), ~~\mathrm{text} \bigg) \bigg],
\label{eq:clip}
\end{equation}
where $\{\mathrm{Proj}_i\}$ is a set of pre-defined projections that uniformly cover different angles and project the point cloud to 2D space. And $S_{\text{clip}}$ is the CLIP loss that calculates the distance between image and text by a pretrained CLIP model, $\mathrm{geneartor}$ is a sampler that transports initial $X_0$ to a meaningful point cloud sample. Note that $\mathrm{generator}(\cdot)$ is fixed. During the optimization, we need to forward then backward through the $\mathrm{generator}$. Thus a multi-step iterative $\mathrm{generator}$ would largely slow down the optimization.



\paragraph{Settings and Results}
Given a text prompt, we optimize Eqn.~\ref{eq:clip} for 100 iterations. 
This amounts to 12 seconds for our method.
In contrast, PVD takes about 15 minutes to generate a text-conditioned sample 
due to its thousands-step sampling. 

In Figure~\ref{fig:textguide}, 
we show that both the shape and category of our generated point clouds correlated 
well with the provided text prompts. It can correctly match the text prompt about the object class as well as the basic properties.


\subsection{Point Cloud Completion}

We further apply our method for point cloud completion in both synthetic and real-world settings. We follow the PVD setup and treat the partial point cloud as the conditional input of the generative model. 
Our goal is to sample meaningful point clouds conditioned on a partial point cloud as input. 
Please refer Appendix for a detailed discussion on our settings. 

\begin{figure}[t]
    \centering
    \includegraphics[width=0.46\textwidth]{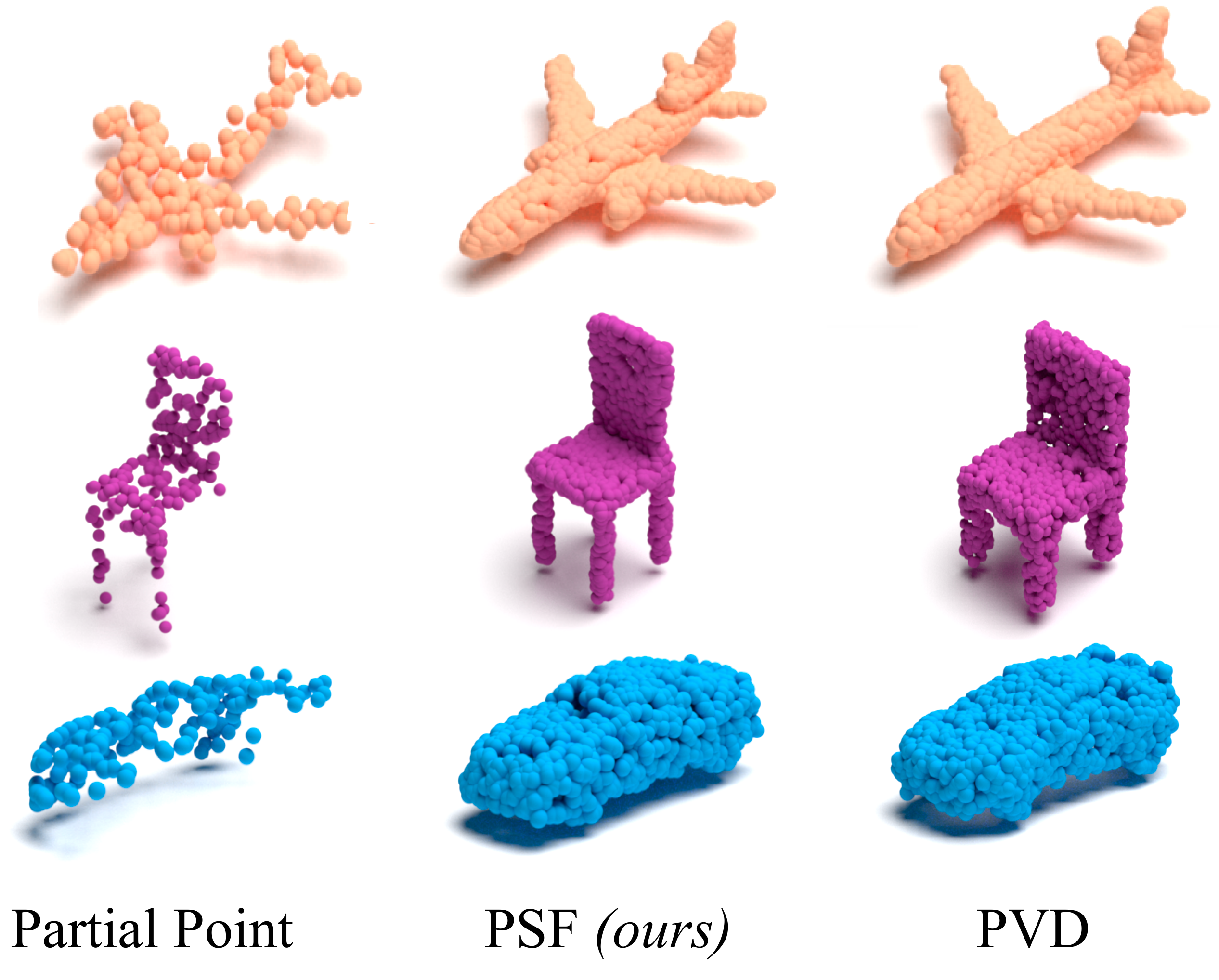}
    \caption{Point Cloud Completion visualization. We show that given the same partial point cloud, we can construct a similar quality point cloud as PVD.}
    \label{fig:compeletion}
\end{figure}

\paragraph{Settings}
We follow PVD and GenRe~\cite{zhang2018learning} and use 20-view depth images rendered from \emph{Chair}, \emph{Airplane}, and \emph{Car} as the input. We sample 200 points as the partial point cloud input for each depth image to train our PSF. 
We EMD as our evaluation metrics as PVD shows that EMD is better than metric compared with CD for the shape completion task.

\paragraph{Results on synthetic shapes}

We summarize our findings in Table~\ref{tab:comp}. 
Our method generates a completed point cloud of similar quality compared to the results from PVD while reducing the completion latency to only $0.04$s.
In Figure~\ref{fig:compeletion}, we show additional qualitative comparisons with PVD. There is no quality degradation visually between our method vs. PVD, further confirming the efficacy of our method on generating high-quality 3D point clouds while being real-time.

\begin{table}[!bpht]
\centering

\begin{tabular}{c|l|ccc}
 \Xhline{3.0\arrayrulewidth} 
Category  & Model & Time (s) $\downarrow$ &    EMD  $\downarrow$ \\
 \Xhline{1.5\arrayrulewidth} 
\multirow{5}{*}{Airplane} 
                  &   SoftFlow~\cite{kim2020softflow}  & 0.12  &  1.198  \\  
                  &   PointFlow~\cite{yang2019pointflow}  & 0.27 &  1.180  \\
                  &   DPF-Net~\cite{klokov2020discrete} & 0.34 &  1.105  \\  
                  &  PVD (N=1000) & 29.98  &  1.030  \\
                  &   PSF (\textit{ours})   &  \textbf{0.04} &  \textbf{1.004}\\
 \Xhline{1.5\arrayrulewidth} 
\multirow{5}{*}{Chair}  
                  &   SoftFlow~\cite{kim2020softflow} & 0.12  &    3.295   \\ 
                  &   PointFlow~\cite{yang2019pointflow}  & 0.27 &  3.649   \\ 
                  &   DPF-Net~\cite{klokov2020discrete}  & 0.34   &   3.320   \\  
                  &   PVD (N=1000) & 29.98   &   2.939\\
                    &   PSF (\textit{ours})  &  \textbf{0.04} &   \textbf{2.937}\\
 \Xhline{1.5\arrayrulewidth} 
\multirow{5}{*}{Car} 
                  &   SoftFlow~\cite{kim2020softflow}  & 0.12 &   2.789  \\ 
                  &   PointFlow~\cite{yang2019pointflow}  & 0.27 &  2.851 \\  
                  &   DPF-Net~\cite{klokov2020discrete}  & 0.34 &   2.318  \\ 
                   &   PVD (N=1000) & 29.98  &   \textbf{2.146}\\
                   &   PSF (\textit{ours}) &  \textbf{0.04} & 2.194\\
 \Xhline{3.0\arrayrulewidth} 
\end{tabular}
\caption{Generate quality and latency between PSF and baselines. CD is multiplied by $10^3$ and EMD is multiplied by  $10^2$. }
\label{tab:comp}
\end{table}

\begin{figure*}[!bhpt]
    \centering
    \includegraphics[width=1.0\textwidth]{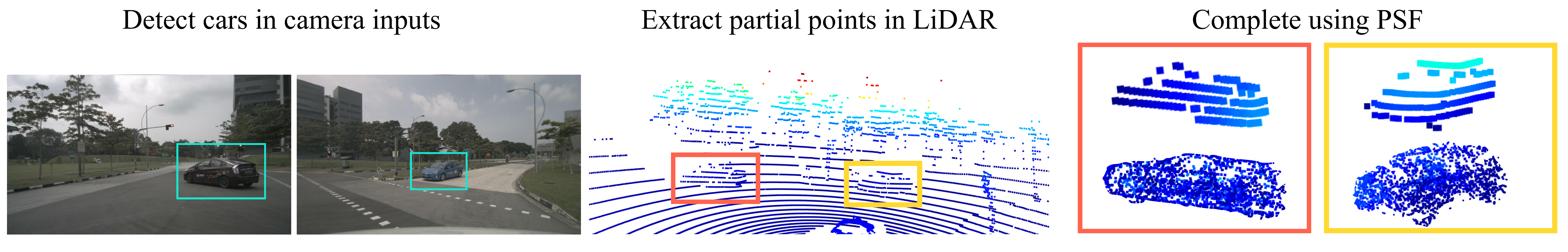}
    \caption{We adapt the PSF in real-world point cloud completion procedure in a low-latency pipeline. All images are from nuScenes~\cite{nuscenes}.}
    \label{fig:car}
\end{figure*}

\begin{figure*}[!bhpt]
    \centering
    \includegraphics[width=1.0\textwidth]{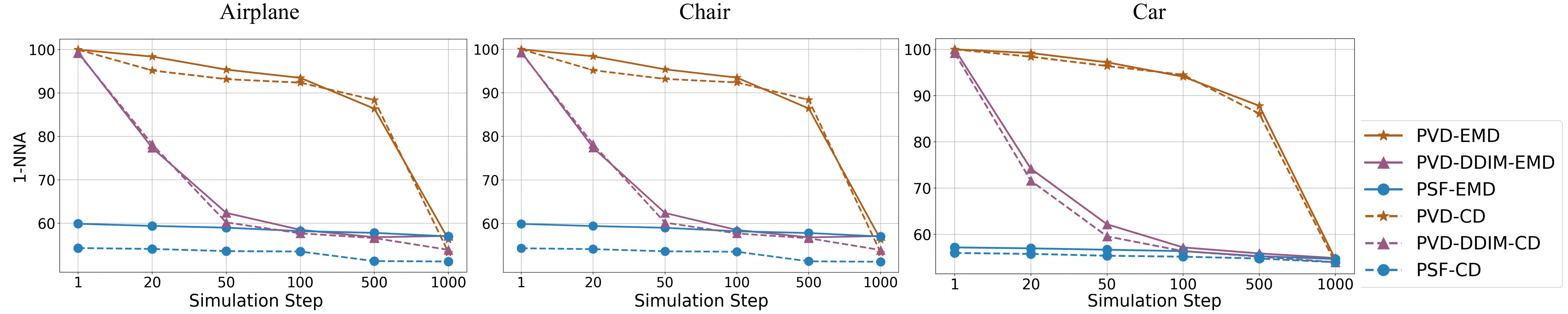}
    \caption{Ablation study on sampling in different steps for PVD, PVD-DDIM and PSF. }
    \label{fig:ablation}
\end{figure*}

\begin{table}[!bhpt]
    \centering
    \vspace{+15pt}
\setlength{\tabcolsep}{1.2mm}
    \renewcommand\arraystretch{1.1}
     \begin{tabular}{l|c|cc|cc|cc}
 \Xhline{3.0\arrayrulewidth} 
       Method & Dist &   \multicolumn{2}{c|}{Airplane} & \multicolumn{2}{c|}{Chair}  & \multicolumn{2}{c}{Car} \\
      & & CD  & EMD  & CD & EMD & CD & EMD \\
 \Xhline{1.5\arrayrulewidth} 
         PVD-DDIM & $\ell_2$ & 85.5 & 83.1 & 82.6 & 79.3 & 80.1 & 77.9 \\
          PVD-DDIM &  CD &  79.9 & 73.1 & 72.4 & 68.2 &  66.3 & 65.7 \\   \Xhline{1.5\arrayrulewidth} 
          PSF \textit{(ours)} &  $\ell_2$  & 79.5 & 73.5 & 68.4 & 62.0 & 67.8 & 67.0 \\
         PSF \textit{(ours)}& CD & \textbf{71.1} & \textbf{65.0} & \textbf{59.9} & \textbf{54.3}& \textbf{57.1} & \textbf{56.0} \\
 \Xhline{3.0\arrayrulewidth} 
    \end{tabular}
    \caption{Ablation study on distillation loss configurations in 1-NNA ($\downarrow$), the CD represents using Chamfer distance as the distance function of the distillation loss. }
    \label{tab:ablation_distill}
    \vspace{-1em}
\end{table}

\paragraph{Transfer to Lidar point cloud completion}
With the fast and high-quality completion result, we are able to apply PSF completion to real-world applications that require low latency. 
One of the important areas that the fast completion benefits are the outdoor 3D detection for autonomous driving.
Usually, the LiDAR scan on the car can only capture a partial view of the sparse point cloud. 
This makes learning a reasonable 3D detector challenging. MVP~\cite{yin2021multimodal} shows that completing and densifying the sparse partial point cloud using nearest neighbor retrieval can lead to better 3D detection performance. To further enhance the completion and densification qualities, most of the current methods, including PVD, are too slow to meet the low latency requirement. To this end, PSF can perform a suitable role by applying this method to enhance the LiDAR scan.


To demonstrate the possibility of our method to enable point cloud completion in low latency, 
in Figure~\ref{fig:car},  we follow MVP and first detect the cars using the camera input with a 2D detector and locate the corresponding partial point clouds in the LiDAR scan. We then apply our model pretrained on \emph{Car} to complete the partial point cloud.
In this example, images and sparse point cloud scans are randomly sampled from nuScenes~\cite{nuscenes}.

Overall, our PSF can generate the completed point clouds for multiple cars in 0.2s, compared with 2D detector and 3D detector that usually cost more than 0.8s for a single scene, our PSF completions is efficient enough for a low-latency requirement.
On the contrary, PVD more than one minute to complete the point clouds.

\subsection{Ablation Study}
\label{sec:ablation}
\paragraph{Different simulation steps}
We consolidate the sampling steps to one step by default.
In this part, 
we further study the performance of the multi-step model without distillation.
We follow the same training settings in Section~\ref{exp:uncondition}. We only apply distillation on one-step and for other steps, we only perform reflow process. We see that reflow model with straight line already gets a well performed model with few steps and the distillation mainly works for pushing the one-step to a similar performance.

\vspace{-10pt}
\paragraph{Distillation loss}
Different from the image representation, which has a well-alignment pixel grid, the point cloud mapping from two distributions is randomly permutated.
We study the impact of different distillation loss choices to show how Chamfer distance better deals with the point cloud representation in distillation setup and how the straightness benefits the distillation. 
From Table~\ref{tab:ablation_distill}, we show that Chamfer distance significantly outperforms the naive $\ell_2$ loss for distillation. 

\section{Discussion and Conclusion}
In this paper, we present a fast point cloud generator, Point Straight Flow, which generates high-quality samples from noise in one step by optimizing the curvy learning transport trajectory. 
Extensive experiments on several standard 3D point cloud benchmarks and real-world applications, including unconditional generation, training-free text-guided generation, and point cloud completion consistently validate PSF's advantages. We also demonstrate that PSF generates a high-fidelity 3D point cloud sample much faster than PVD and the PVD-DDIM. 

Our method is easy to formulate and implement, which can serve as an alternative to standard diffusion-based point cloud generator.
Our preliminary work suggests that PSF has potential applications beyond complex scene completion and text-guided generation applications.

\clearpage
\newpage

{\small
\bibliographystyle{ieeefullname}
\bibliography{reference}

\begin{thebibliography}{10}\itemsep=-1pt

\bibitem{achlioptas2018learning}
Panos Achlioptas, Olga Diamanti, Ioannis Mitliagkas, and Leonidas Guibas.
\newblock Learning representations and generative models for 3d point clouds.
\newblock In {\em International conference on machine learning}, pages 40--49.
  PMLR, 2018.

\bibitem{albergo2022building}
Michael~S Albergo and Eric Vanden-Eijnden.
\newblock Building normalizing flows with stochastic interpolants.
\newblock {\em arXiv preprint arXiv:2209.15571}, 2022.

\bibitem{brock2016generative}
Andrew Brock, Theodore Lim, James~M Ritchie, and Nick Weston.
\newblock Generative and discriminative voxel modeling with convolutional
  neural networks.
\newblock {\em arXiv preprint arXiv:1608.04236}, 2016.

\bibitem{nuscenes}
Holger Caesar, Varun Bankiti, Alex~H. Lang, Sourabh Vora, Venice~Erin Liong,
  Qiang Xu, Anush Krishnan, Yu Pan, Giancarlo Baldan, and Oscar Beijbom.
\newblock nuscenes: A multimodal dataset for autonomous driving.
\newblock In {\em CVPR}, 2020.

\bibitem{cai2020learning}
Ruojin Cai, Guandao Yang, Hadar Averbuch-Elor, Zekun Hao, Serge Belongie, Noah
  Snavely, and Bharath Hariharan.
\newblock Learning gradient fields for shape generation.
\newblock In {\em European Conference on Computer Vision}, pages 364--381.
  Springer, 2020.

\bibitem{chen2018neural}
Ricky~TQ Chen, Yulia Rubanova, Jesse Bettencourt, and David~K Duvenaud.
\newblock Neural ordinary differential equations.
\newblock {\em Advances in neural information processing systems}, 31, 2018.

\bibitem{dhariwal2021diffusion}
Prafulla Dhariwal and Alexander Nichol.
\newblock Diffusion models beat gans on image synthesis.
\newblock {\em Advances in Neural Information Processing Systems},
  34:8780--8794, 2021.

\bibitem{gao2022get3d}
Jun Gao, Tianchang Shen, Zian Wang, Wenzheng Chen, Kangxue Yin, Daiqing Li, Or
  Litany, Zan Gojcic, and Sanja Fidler.
\newblock Get3d: A generative model of high quality 3d textured shapes learned
  from images.
\newblock In {\em Advances In Neural Information Processing Systems}, 2022.

\bibitem{goodfellow2014generative}
Ian Goodfellow, Jean Pouget-Abadie, Mehdi Mirza, Bing Xu, David Warde-Farley,
  Sherjil Ozair, Aaron Courville, and Yoshua Bengio.
\newblock Generative adversarial nets.
\newblock {\em Advances in neural information processing systems}, 27, 2014.

\bibitem{ho2020denoising}
Jonathan Ho, Ajay Jain, and Pieter Abbeel.
\newblock Denoising diffusion probabilistic models.
\newblock {\em Advances in Neural Information Processing Systems},
  33:6840--6851, 2020.

\bibitem{ho2022video}
Jonathan Ho, Tim Salimans, Alexey Gritsenko, William Chan, Mohammad Norouzi,
  and David~J Fleet.
\newblock Video diffusion models.
\newblock {\em arXiv preprint arXiv:2204.03458}, 2022.

\bibitem{kim2020softflow}
Hyeongju Kim, Hyeonseung Lee, Woo~Hyun Kang, Joun~Yeop Lee, and Nam~Soo Kim.
\newblock Softflow: Probabilistic framework for normalizing flow on manifolds.
\newblock {\em Advances in Neural Information Processing Systems},
  33:16388--16397, 2020.

\bibitem{kim2021setvae}
Jinwoo Kim, Jaehoon Yoo, Juho Lee, and Seunghoon Hong.
\newblock Setvae: Learning hierarchical composition for generative modeling of
  set-structured data.
\newblock In {\em Proceedings of the IEEE/CVF Conference on Computer Vision and
  Pattern Recognition}, pages 15059--15068, 2021.

\bibitem{kingma2013auto}
Diederik~P Kingma and Max Welling.
\newblock Auto-encoding variational bayes.
\newblock {\em arXiv preprint arXiv:1312.6114}, 2013.

\bibitem{klokov2020discrete}
Roman Klokov, Edmond Boyer, and Jakob Verbeek.
\newblock Discrete point flow networks for efficient point cloud generation.
\newblock In {\em European Conference on Computer Vision}, pages 694--710.
  Springer, 2020.

\bibitem{kong2021fast}
Zhifeng Kong and Wei Ping.
\newblock On fast sampling of diffusion probabilistic models.
\newblock {\em arXiv preprint arXiv:2106.00132}, 2021.

\bibitem{li2021sp}
Ruihui Li, Xianzhi Li, Ka-Hei Hui, and Chi-Wing Fu.
\newblock Sp-gan: Sphere-guided 3d shape generation and manipulation.
\newblock {\em ACM Transactions on Graphics (TOG)}, 40(4):1--12, 2021.

\bibitem{lipman2022flow}
Yaron Lipman, Ricky~TQ Chen, Heli Ben-Hamu, Maximilian Nickel, and Matt Le.
\newblock Flow matching for generative modeling.
\newblock {\em arXiv preprint arXiv:2210.02747}, 2022.

\bibitem{liu2022rectified}
Qiang Liu.
\newblock Rectified flow: A marginal preserving approach to optimal transport.
\newblock {\em arXiv preprint arXiv:2209.14577}, 2022.

\bibitem{liu2022flow}
Xingchao Liu, Chengyue Gong, and Qiang Liu.
\newblock Flow straight and fast: Learning to generate and transfer data with
  rectified flow.
\newblock {\em arXiv preprint arXiv:2209.03003}, 2022.

\bibitem{liu2021fusedream}
Xingchao Liu, Chengyue Gong, Lemeng Wu, Shujian Zhang, Hao Su, and Qiang Liu.
\newblock Fusedream: Training-free text-to-image generation with improved clip+
  gan space optimization.
\newblock {\em arXiv preprint arXiv:2112.01573}, 2021.

\bibitem{pvcnn}
Zhijian Liu, Haotian Tang, Yujun Lin, and Song Han.
\newblock Point-voxel cnn for efficient 3d deep learning.
\newblock {\em Advances in Neural Information Processing Systems}, 32, 2019.

\bibitem{luhman2021knowledge}
Eric Luhman and Troy Luhman.
\newblock Knowledge distillation in iterative generative models for improved
  sampling speed.
\newblock {\em arXiv preprint arXiv:2101.02388}, 2021.

\bibitem{luo2021diffusion}
Shitong Luo and Wei Hu.
\newblock Diffusion probabilistic models for 3d point cloud generation.
\newblock In {\em Proceedings of the IEEE/CVF Conference on Computer Vision and
  Pattern Recognition}, pages 2837--2845, 2021.

\bibitem{lyu2021conditional}
Zhaoyang Lyu, Zhifeng Kong, Xudong Xu, Liang Pan, and Dahua Lin.
\newblock A conditional point diffusion-refinement paradigm for 3d point cloud
  completion.
\newblock {\em arXiv preprint arXiv:2112.03530}, 2021.

\bibitem{michel2022text2mesh}
Oscar Michel, Roi Bar-On, Richard Liu, Sagie Benaim, and Rana Hanocka.
\newblock Text2mesh: Text-driven neural stylization for meshes.
\newblock In {\em Proceedings of the IEEE/CVF Conference on Computer Vision and
  Pattern Recognition}, pages 13492--13502, 2022.

\bibitem{nichol2021improved}
Alexander~Quinn Nichol and Prafulla Dhariwal.
\newblock Improved denoising diffusion probabilistic models.
\newblock In {\em International Conference on Machine Learning}, pages
  8162--8171. PMLR, 2021.

\bibitem{papamakarios2021normalizing}
George Papamakarios, Eric~T Nalisnick, Danilo~Jimenez Rezende, Shakir Mohamed,
  and Balaji Lakshminarayanan.
\newblock Normalizing flows for probabilistic modeling and inference.
\newblock {\em J. Mach. Learn. Res.}, 22(57):1--64, 2021.

\bibitem{ramesh2022hierarchical}
Aditya Ramesh, Prafulla Dhariwal, Alex Nichol, Casey Chu, and Mark Chen.
\newblock Hierarchical text-conditional image generation with clip latents.
\newblock {\em arXiv preprint arXiv:2204.06125}, 2022.

\bibitem{rombach2022high}
Robin Rombach, Andreas Blattmann, Dominik Lorenz, Patrick Esser, and Bj{\"o}rn
  Ommer.
\newblock High-resolution image synthesis with latent diffusion models.
\newblock In {\em Proceedings of the IEEE/CVF Conference on Computer Vision and
  Pattern Recognition}, pages 10684--10695, 2022.

\bibitem{ronneberger2015u}
Olaf Ronneberger, Philipp Fischer, and Thomas Brox.
\newblock U-net: Convolutional networks for biomedical image segmentation.
\newblock In {\em International Conference on Medical image computing and
  computer-assisted intervention}, pages 234--241. Springer, 2015.

\bibitem{salimans2022progressive}
Tim Salimans and Jonathan Ho.
\newblock Progressive distillation for fast sampling of diffusion models.
\newblock {\em arXiv preprint arXiv:2202.00512}, 2022.

\bibitem{shu20193d}
Dong~Wook Shu, Sung~Woo Park, and Junseok Kwon.
\newblock 3d point cloud generative adversarial network based on tree
  structured graph convolutions.
\newblock In {\em Proceedings of the IEEE/CVF international conference on
  computer vision}, pages 3859--3868, 2019.

\bibitem{song2020denoising}
Jiaming Song, Chenlin Meng, and Stefano Ermon.
\newblock Denoising diffusion implicit models.
\newblock {\em arXiv preprint arXiv:2010.02502}, 2020.

\bibitem{song2019generative}
Yang Song and Stefano Ermon.
\newblock Generative modeling by estimating gradients of the data distribution.
\newblock {\em Advances in Neural Information Processing Systems}, 32, 2019.

\bibitem{song2020score}
Yang Song, Jascha Sohl-Dickstein, Diederik~P Kingma, Abhishek Kumar, Stefano
  Ermon, and Ben Poole.
\newblock Score-based generative modeling through stochastic differential
  equations.
\newblock In {\em International Conference on Learning Representations}, 2020.

\bibitem{tzen2019theoretical}
Belinda Tzen and Maxim Raginsky.
\newblock Theoretical guarantees for sampling and inference in generative
  models with latent diffusions.
\newblock In {\em Conference on Learning Theory}, pages 3084--3114. PMLR, 2019.

\bibitem{3dgan}
Jiajun Wu, Chengkai Zhang, Tianfan Xue, William~T Freeman, and Joshua~B
  Tenenbaum.
\newblock Learning a probabilistic latent space of object shapes via 3d
  generative-adversarial modeling.
\newblock In {\em Advances in Neural Information Processing Systems}, pages
  82--90, 2016.

\bibitem{yang2019pointflow}
Guandao Yang, Xun Huang, Zekun Hao, Ming-Yu Liu, Serge Belongie, and Bharath
  Hariharan.
\newblock Pointflow: 3d point cloud generation with continuous normalizing
  flows.
\newblock In {\em Proceedings of the IEEE/CVF International Conference on
  Computer Vision}, pages 4541--4550, 2019.

\bibitem{yang2022diffusion}
Ruihan Yang, Prakhar Srivastava, and Stephan Mandt.
\newblock Diffusion probabilistic modeling for video generation.
\newblock {\em arXiv preprint arXiv:2203.09481}, 2022.

\bibitem{yin2021multimodal}
Tianwei Yin, Xingyi Zhou, and Philipp Kr{\"a}henb{\"u}hl.
\newblock Multimodal virtual point 3d detection.
\newblock {\em NeurIPS}, 2021.

\bibitem{zeng2022lion}
Xiaohui Zeng, Arash Vahdat, Francis Williams, Zan Gojcic, Or Litany, Sanja
  Fidler, and Karsten Kreis.
\newblock Lion: Latent point diffusion models for 3d shape generation.
\newblock In {\em Advances in Neural Information Processing Systems (NeurIPS)},
  2022.

\bibitem{zhang2018learning}
Xiuming Zhang, Zhoutong Zhang, Chengkai Zhang, Josh Tenenbaum, Bill Freeman,
  and Jiajun Wu.
\newblock Learning to reconstruct shapes from unseen classes.
\newblock {\em Advances in neural information processing systems}, 31, 2018.

\bibitem{zheng2022neural}
Yan Zheng, Lemeng Wu, Xingchao Liu, Zhen Chen, Qiang Liu, and Qixing Huang.
\newblock Neural volumetric mesh generator.
\newblock {\em arXiv preprint arXiv:2210.03158}, 2022.

\bibitem{zhou20213d}
Linqi Zhou, Yilun Du, and Jiajun Wu.
\newblock 3d shape generation and completion through point-voxel diffusion.
\newblock In {\em Proceedings of the IEEE/CVF International Conference on
  Computer Vision}, pages 5826--5835, 2021.

\end{thebibliography}
}
\clearpage
\pagenumbering{arabic} 
\newpage
\appendix
\section*{Appendix}
This appendix presents a pseudo-code implementation of our Point Straight Flow (PSF), provides experimental training details on point cloud completion, and describes further experimental results on 3D point cloud applications.  
\section{Method}
\subsection{Pseudo-code implementation of PSF}
Our PSF has a simple formulation and is easy to implement with impressive performance on 3D point cloud generation. In Algorithm 2-4, we present a pseudo-code implementation for each stage of our proposed PSF.
\begin{algorithm}

\caption{Train velocity flow model} \label{alg:s1}
\begin{algorithmic}
\STATE \textbf{Input}: Point cloud dataset $\mathcal{D}$.
\STATE \textbf{Input}: Neural velocity field $v_\theta$ with parameter $\theta$.
\STATE
\STATE ~~~ \textbf{for} $K_{\text{train}}$ steps \textbf{do}
\STATE ~~~~~~ {\texttt{\# Construct Intermediate data}}
\STATE ~~~~~~ $X_1 \sim \mathcal{D}$
\STATE ~~~~~~ $X_0 \sim \mathcal{N}(\textbf{0}, \textbf{I})$
\STATE ~~~~~~ $t \sim \mathcal{U}(0, 1)$
\STATE ~~~~~~ $X_t = t X_1 + (1-t) X_0$
\STATE ~~~~~~ $L_{\theta} = \|(v_\theta(X_t, t) - (X_1 - X_0)\|^2$

\STATE ~~~~~~ $\theta \leftarrow \theta - \gamma_{\text{train}} \nabla_{\theta}L_{\theta}$
\STATE ~~~ \textbf{done}
\STATE
\STATE \textbf{Output}: Trained network $v_\theta$ with parameter $\theta$.

\end{algorithmic}

\end{algorithm}

\begin{algorithm}

\caption{Improving straightness via \textit{reflow}} \label{alg:s2}
\begin{algorithmic}
\STATE \textbf{Input}: Pretrained neural velocity field $v_\theta$ in Algorithm~\ref{alg:s1}.
\STATE ~~~ {\texttt{\# Sample pairs set $\mathcal{S} = \{(X'_0, X'_1)\}$}}
\STATE ~~~ \textbf{for} $K_{\textit{sample}}$ steps \textbf{do}

\STATE ~~~~~~ $X'_0 \sim \mathcal{N}(\textbf{0}, \textbf{I})$
\STATE ~~~~~~ for $\hat{t}$ in $\{0, 1, ..., N-1\}$ do
\STATE ~~~~~~~~~  $X'_{(\hat{t}+1)/N} \longleftarrow X'_{\hat{t}/N} + \frac{1}{N} ~v_\theta(X'_{\hat{t}}, \frac{\hat{t}}{N})$
\STATE ~~~~~~ \textbf{done}
\STATE ~~~~~~ $X'_1 := X'_{(N-1+1)/N}$
\STATE ~~~~~~ Add pair $(X_0, X_1)$ into $\mathcal{S}$
\STATE ~~~ \textbf{done}

\STATE

\STATE ~~~ {\texttt{\# \textit{reflow} procedure}}
\STATE ~~~ \textbf{for} $K_{\textit{reflow}}$ steps \textbf{do}

\STATE ~~~~~~ $(X'_0, X'_1) \sim \mathcal{S}$
\STATE ~~~~~~ $t \sim \mathcal{U}(0, 1)$
\STATE ~~~~~~ $X'_t = t X'_1 + (1-t) X'_0$
\STATE ~~~~~~ $L_{\theta} = \|(v_\theta(X_t, t) - (X_1 - X_0)\|^2$
\STATE ~~~~~~ $\theta \leftarrow \theta - \gamma_{\text{reflow}} \nabla_{\theta}L_{\theta}$
\STATE ~~~ \textbf{done}

\STATE \textbf{Output}: Finetuned network $v_\theta$ with parameter $\theta$.

\end{algorithmic}

\end{algorithm}

\begin{algorithm}

\caption{Flow distillation} \label{alg:s3}
\begin{algorithmic}
\STATE \textbf{Input}: Finetuned neural velocity field $v_\theta$ in Algorithm~\ref{alg:s2}.
~~~ \textbf{Input}: Sampled data pairs set $\mathcal{S}$.
\STATE ~~~ {\texttt{\# Distill for a one-step model}}
\STATE ~~~ \textbf{for} $K_{\textit{distill}}$ steps \textbf{do}

\STATE ~~~~~~ $(X'_0, X'_1) \sim \mathcal{S}$
\STATE ~~~~~~ $L_{\theta} = \text{CD}(v_\theta(X_0, 0) + X_0, X_1)$

\STATE ~~~~~~ $\theta \leftarrow \theta - \gamma_{\text{distill}} \nabla_{\theta}L_{\theta}$
\STATE ~~~ \textbf{done}

\STATE \textbf{Output}: Distilled network $v_\theta$ with parameter $\theta$ as the final PSF.

\end{algorithmic}

\end{algorithm}

The hyperparameter $\gamma$ in Algorithm 2 denotes the learning rate. For each stage in the experiment, described in Section~\ref{sec:exp}.

\subsection{Details of point cloud completion setup}
Follow PVD~\cite{zhou20213d}, we extend the application of our  model from unconditional generation to the conditional shape completion. The general setup of the shape completion is to complete the rest part of the shape with a given partial point cloud input. Let us take $c\in \mathbb{R}^{M \times 3}$ to  describe the partial point cloud input, and use neural network to predict the drift force with the conditional input. That means we are only required to change the drift force term into

\begin{equation}
    v_{\theta}(X_t, t, c),
\end{equation}

Prior to the \textit{reflow} and distillation procedure, we additionally record the randomly sampled partial point cloud $c$ together with random sampled $X'_0$ and the corresponding generated $X'_1$ as the fintuning data in $(X'_0, X'_1, c)$.
\begin{table*}[!bhpt]
\centering
\small
\begin{tabular}{l|cc|cc|cc|cc|cc|cc}
 \Xhline{3.0\arrayrulewidth} 
Class &
& \multicolumn{2}{c}{Airplane} & \multicolumn{6}{c}{Chair} & \multicolumn{2}{c}{Car} \\
\hline
& \multicolumn{2}{c}{MMD$\downarrow$} &
\multicolumn{2}{c}{COV$\uparrow$ } & \multicolumn{2}{c}{MMD$\downarrow$}&
\multicolumn{2}{c}{COV$\uparrow$ } & \multicolumn{2}{c}{MMD$\downarrow$}&
\multicolumn{2}{c}{COV$\uparrow$ } \\ 
\hline
         &    CD   &     EMD &  CD    &     EMD   &    CD   &     EMD &  CD    &     EMD   &    CD   &     EMD &  CD    &     EMD   \\ 
 \Xhline{3.0\arrayrulewidth} 
            
                l-GAN (CD)~\cite{achlioptas2018learning}  & 0.3398 & 0.5832 & 38.52 & 21.23 & 2.589 & 2.007 & 41.99 & 29.31 & 1.532 & 1.226 & 38.92 & 23.58\\
                  PointFlow~\cite{yang2019pointflow}    &0.2243 &0.3901 &47.90 &46.41 & \textbf{2.409}&1.595 &42.90 &50.00 & \textbf{0.9010}& 0.8071& 46.88&50.00 \\ 
                  SoftFlow~\cite{kim2020softflow}     & 0.2309 & 0.3745 &46.91 &47.90 & 2.528 & 1.682 & 41.39 & 47.43 & 1.187 & 0.8594 & 42.90 & 44.60 \\ 
                   DPF-Net~\cite{klokov2020discrete}  &0.2642 &0.4086 &46.17 & 48.89 & 2.536&1.632 &44.71 &48.79 &1.129 & 0.8529& 45.74&49.43 \\ 
                  Shape-GF~\cite{cai2020learning}  & 2.703 & 0.6592 & 40.74 &40.49 & 2.889 & 1.702 & 46.67 & 48.03 & 9.232 & \textbf{0.7558} &\textbf{49.43} & 50.28\\

 PVD &0.2243 &0.3803 &\textbf{48.88} &52.09 &2.622&\textbf{1.556} &\textbf{49.84} & \textbf{50.60} &1.077 & 0.7938 &41.19 & 50.56 \\ 
 \hline
  PVD-DDIM (N=100) & 0.2434 &0.3991 & 44.23 & 49.75 &2.758 & 1.703 & 46.32 & 48.19 &1.202 & 0.8176 &40.01 &48.34 \\ 
   PSF (N=1) \textit{(ours)} &\textbf{0.2205} & \textbf{0.3661} &46.17 &\textbf{52.59} &2.624& 1.573 &46.71 & 49.84 & 1.023 & 0.8020 & 42.89 &\textbf{53.12} \\ 
 \Xhline{3.0\arrayrulewidth} 
\end{tabular}

\caption{Further experimental results on unconditional 3d point cloud generation with MMD and COV scores. The scale is aligned  with PVD~\cite{zhou20213d}.}
\label{tab:gen_full_metrics}
\end{table*}


\begin{figure*}[!bhpt]
    \centering
    \includegraphics[width=0.9\textwidth]{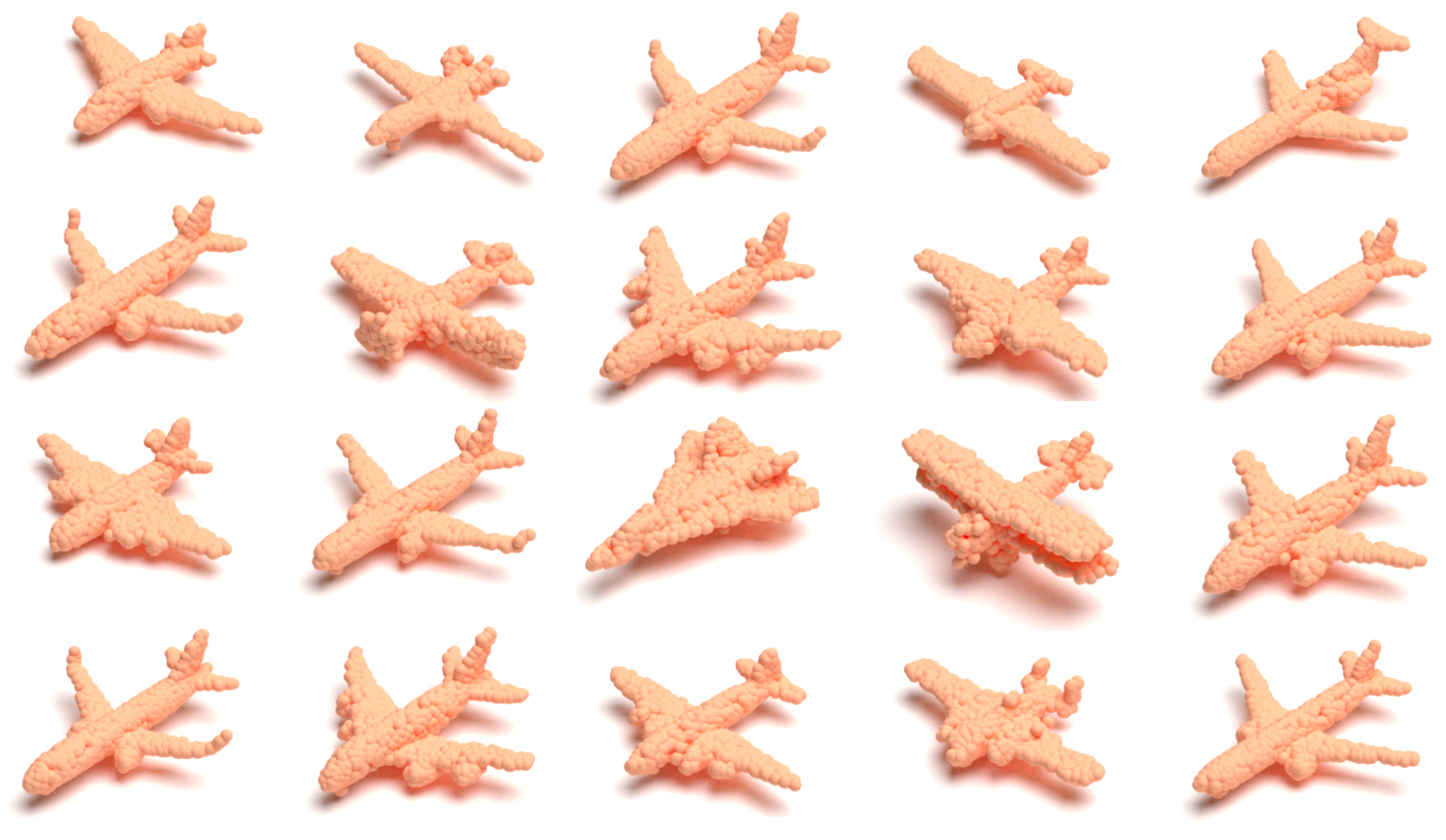}
    \caption{Visualization results of PSF generated Airplane samples.}
    \label{fig:appen_air}
\end{figure*}

\section{Further Experimental Results}

\subsection{Unconditional 3d point cloud generation}

\paragraph{MMD and COV metrics} We report the MMD and COV score in CD and EMD distance for three classes, e.g., Airplane, Chair and Car. We show that PSF with one step can perform comparably or even better in some categories than 1-NNA when comparing to  PVD. The impressive performance is attributed to the ODE fashion transport style as well as the desired distillation properties in PSF.

\paragraph{Reflow result before distillation}

\begin{figure*}[!bhpt]
    \centering
    \includegraphics[width=1.0\textwidth]{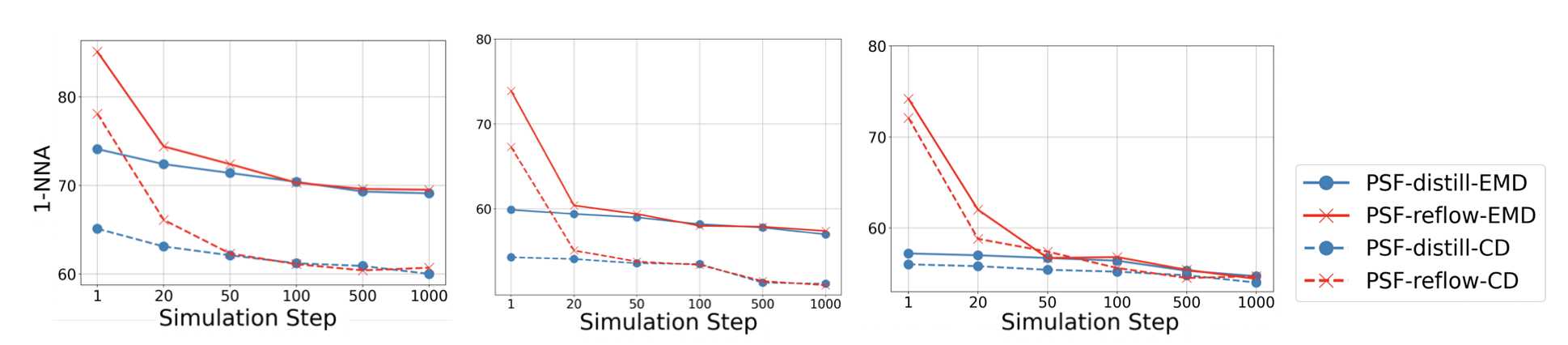}
    \caption{Results of PSF before distillation and after distillation. We show that PSF-reflow achieves similar results as PSF-distill. Distillation is mainly serve as a one-step generation.}
    \label{fig:appen-reflow}
\end{figure*}

We show that the PSF after reflow without distillation can provide smooth shape in a few-step setup. We denote the PSF before the distillation as PSF-reflow and PSF after distillation as PSF-distill. Figure~\ref{fig:appen-reflow} shows that PSF-reflow can obtain similar results as PST-distill when iterating up to 20 steps, which indicates a very straight trajectory after applying reflow procedure. These results further suggest that the distillation is mainly useful in extreme small steps to help correct some inaccurate directions.

\begin{figure*}[!bhpt]
    \centering
    \includegraphics[width=0.9\textwidth]{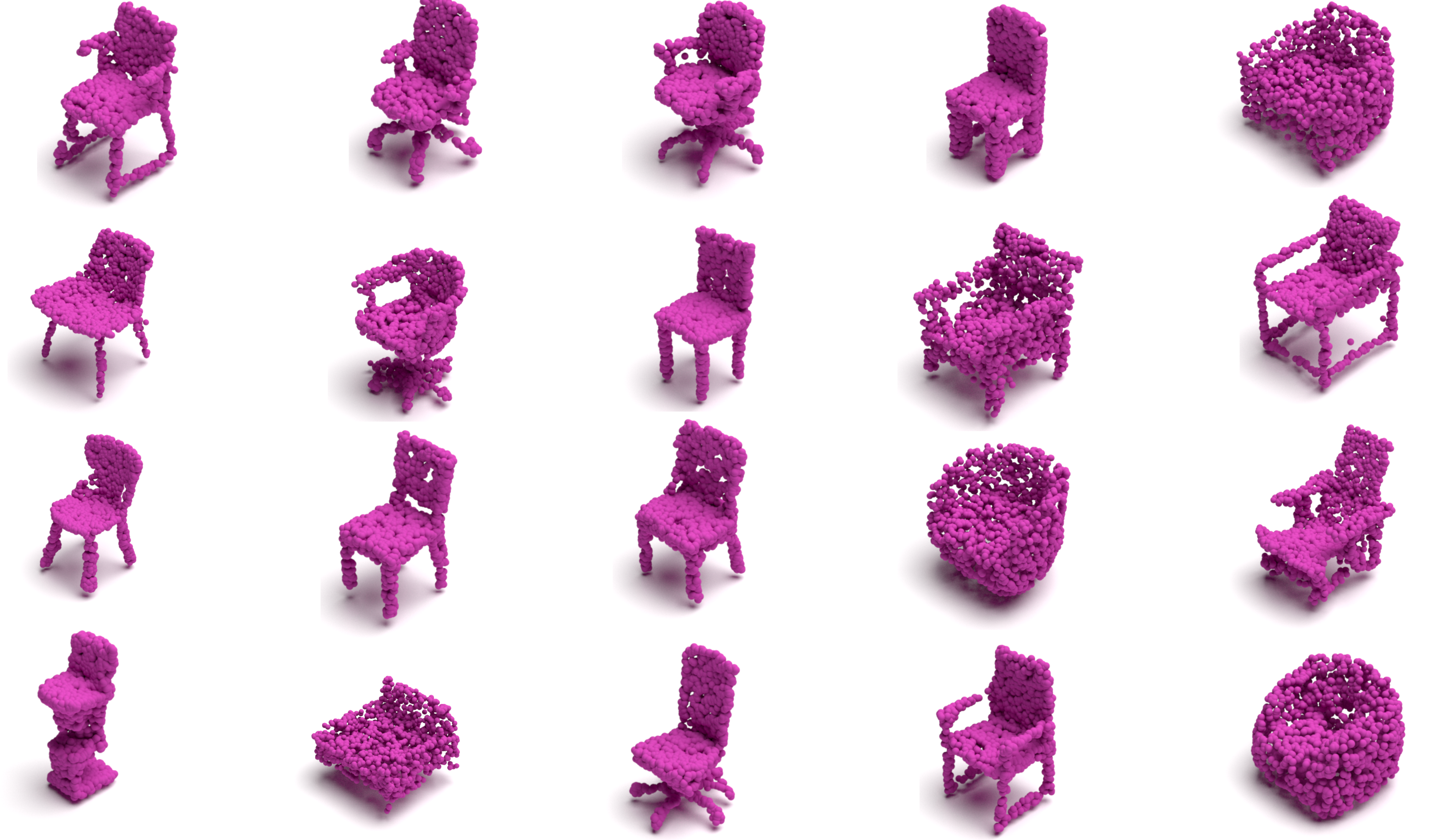}
    \caption{Visualization results of PSF generated Chair samples.}
    \label{fig:appen_chair}
\end{figure*}

\begin{figure*}[!bhpt]
    \centering
    \includegraphics[width=0.9\textwidth]{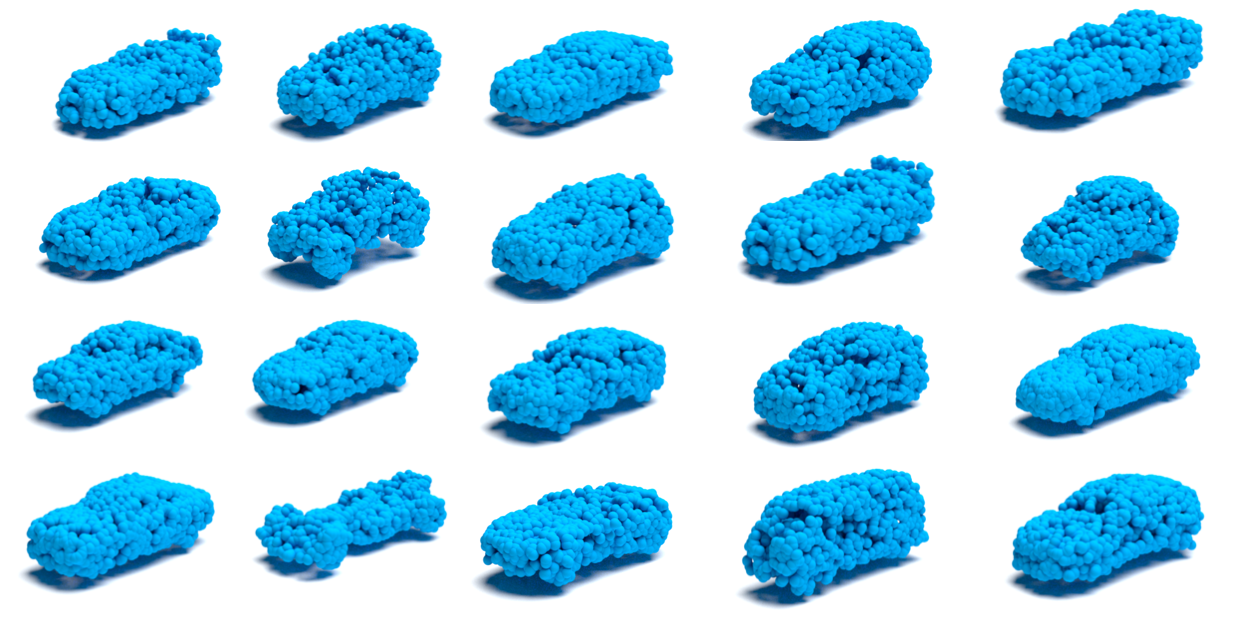}
    \caption{Visualization results of PSF generated Car samples}
    \label{fig:appen_car}
\end{figure*}

\paragraph{More qualitative results}
We show more qualitative results for visualizing generated Airplane, Chair, and Car shapes in Figure~\ref{fig:appen_air}, \ref{fig:appen_chair} and \ref{fig:appen_car} by randomly sampling 20 point clouds without any cherry-picking. These results indicate that PSF can consistently provide reasonable shapes.


\subsection{Point Completion}
\paragraph{Further experimental results on real-world application}

\begin{figure*}[!bhpt]
    \centering
    \includegraphics[width=1.0\textwidth]{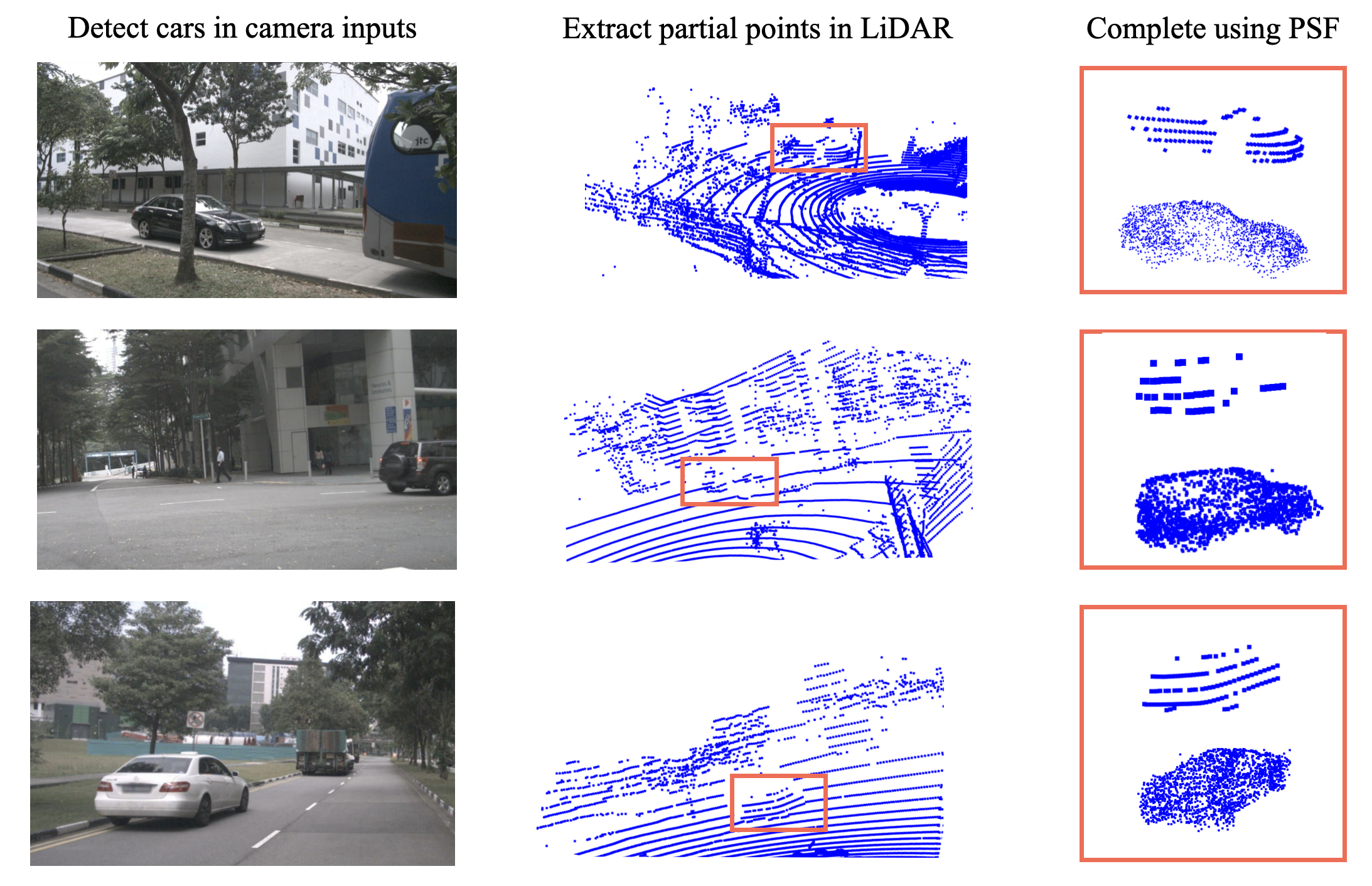}
    \caption{Further completion results on real-world application. Image data is from nuScene~\cite{nuscenes}.}
    \label{fig:appendix_comp}
\end{figure*}

We here show more experimental results on car point cloud completion in Figure~\ref{fig:appendix_comp}. Our preliminary work suggests that our fast PSF can eventually benefit many 3D perceptual tasks in the future.

\end{document}